\documentclass[a4paper,conference]{IEEEtran}
\ifCLASSINFOpdf
\else
\fi

\usepackage[a4paper,
            left=0.514in,
            right=0.514in,
            top=0.75in,
            bottom=1.59in,
            ]{geometry}

\usepackage[inline]{enumitem}
\usepackage{graphicx}
\usepackage{epsfig}
\usepackage[belowskip=0pt]{caption}
\usepackage[belowskip=0pt]{subcaption}
\usepackage{booktabs}
\usepackage{multirow}
\usepackage{hyperref}
\usepackage{multicol}
\usepackage{tabularx}
\usepackage{amsmath}
\usepackage{algorithm}
\usepackage{array}
\usepackage{siunitx}
\usepackage{url}
\usepackage{dblfloatfix} 
\usepackage[nice]{nicefrac}
\usepackage{algpseudocode}

\newcolumntype{P}[1]{>{\centering\arraybackslash}p{#1}}
\newcolumntype{M}[1]{>{\centering\arraybackslash}m{#1}}
\newcommand*{\Comb}[2]{{}^{#1}C_{#2}}%

\begin{document}
%
\title{Transformer based Fingerprint Feature Extraction}

\author{\IEEEauthorblockN{Saraansh Tandon}
\IEEEauthorblockA{International Institute of Information and Technology,\\
Hyderabad, India 500032\\
Email: saraansh.tandon@research.iiit.ac.in}
\and
\IEEEauthorblockN{Anoop Namboodiri}
\IEEEauthorblockA{International Institute of Information and Technology,\\
Hyderabad, India 500032\\
Email: anoop@iiit.ac.in}
}


%


\maketitle

\begin{abstract}
Fingerprint feature extraction is a task that is solved using either a global or a local representation. State-of-the-art global approaches use heavy deep learning models to process the full fingerprint image at once, which makes the corresponding approach memory intensive. On the other hand, local approaches involve minutiae based patch extraction, multiple feature extraction steps and an expensive matching stage, which make the corresponding approach time intensive. However, both these approaches provide useful and sometimes exclusive insights for solving the problem. Using both approaches together for extracting fingerprint representations is semantically useful but quite inefficient. Our convolutional transformer based approach with an in-built minutiae extractor provides a time and memory efficient solution to extract a global as well as a local representation of the fingerprint. The use of these representations along with a smart matching process gives us state-of-the-art performance across multiple databases. The project page can be found at https://saraansh1999.github.io/global-plus-local-fp-transformer.
\end{abstract}


%
\IEEEpeerreviewmaketitle

\section{Introduction}
\label{sec:intro}
From daily usage in smartphones to granting access to highly confidential resources, fingerprints have proven to be a reliable source of biometric identity. Hence it comes as no surprise that this field has been a very active domain of research since many decades.
Most fingerprint feature extraction algorithms today are built upon the domain knowledge of global features like ridge-flow or local features like minutiae and pores.
Knowledge of at least one of these, possibly in combination with others, is used to process an input fingerprint image and output a representation. This representation acts as a proxy of the image and is used to determine the identity of the owner of the fingerprint by matching it with a database.
\\
\textbf{Global approaches} for extracting fixed-length fingerprint representations function by processing the full image as shown in Figure \ref{fig:global_intro}. One of the first such attempts was made by Jain et al. in Fingercode \cite{DBLP:conf/cvpr/JainPHP99} where a set of Gabor filters were used to produce a 640 byte long representation. Later as the deep learning era emerged, \cite{DBLP:conf/icb/CaoJ17} and \cite{SongAndFeng} made use of convolutional networks for the same. Recently, Engelsma et al. proposed DeepPrint\cite{DBLP:journals/pami/EngelsmaCJ21} which uses a multi-task approach to incorporate domain knowledge into deep learning of fingerprint representations and obtains state-of-the-art performance among fixed-length global approaches.
\\
\textbf{Local approaches} function by breaking down the input image into multiple local patches based on some local features. The most popular of these local features of interest are the minutiae points, because of their studied permanence \cite{yoon2015} and uniqueness \cite{pakanti2002} among humans. \cite{MCC}, \cite{SONG2019397}, and \cite{Li2019LearningGF} proposed methods that led to fixed-length representations,
but local approaches may also lead to variable-length representations whose size depends on the number of minutiae identified in the input image.  Cao et al. proposed such an end-to-end method called LatentAFIS\cite{latentafis} as shown in Figure \ref{fig:local_intro}. It treats the entire set of the predicted minutiae and their corresponding embeddings as the representation of the full fingerprint. 
\\
{
Since DeepPrint\cite{DBLP:journals/pami/EngelsmaCJ21} and LatentAFIS\cite{latentafis} obtain state-of-the-art performances in their respective domains, we treat them as baselines for our approach. On analyzing the mutually exclusive failure cases of \cite{DBLP:journals/pami/EngelsmaCJ21} and \cite{latentafis} we observe that the global and local perspectives can compensate for the semantic shortcomings of each other. Figure \ref{fig:fails_dp} shows a pair of fingerprints originating from two different identities, but due to a similar ridge-flow structure, \cite{DBLP:journals/pami/EngelsmaCJ21} marks it as a true pair. Whereas \cite{latentafis} is able to focus on the subtle differences existing in the minutiae and mark it accurately as a false pair. Figure \ref{fig:fails_la} shows a pair of impressions from the same finger, but \cite{popli2021unified} claims that distortion is able to fool \cite{latentafis} which misses a lot of minutiae matches as the corresponding patches look significantly different. \cite{DBLP:journals/pami/EngelsmaCJ21} on the other hand is able to look at the big picture similarities at the center of the fingerprint to make the correct prediction. Note that \cite{DBLP:journals/pami/EngelsmaCJ21} is inferior in overall performance as compared to \cite{latentafis} as the latter is much more strongly tied to the domain knowledge of fingerprints.
\begin{figure}[h]
    \centering
    \begin{subfigure}[t]{0.45\columnwidth}
        \centering
        \includegraphics[width=\columnwidth]{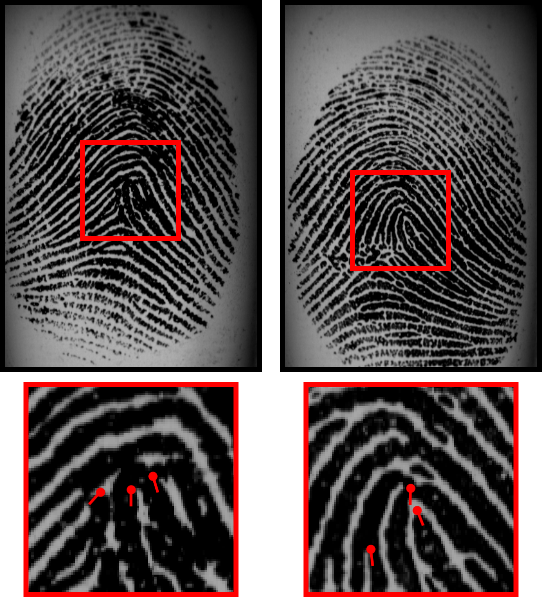}
        \caption{}
        \label{fig:fails_dp}
    \end{subfigure}
    \hspace{1em}
    \begin{subfigure}[t]{0.45\columnwidth}
        \centering
        \includegraphics[width=\columnwidth]{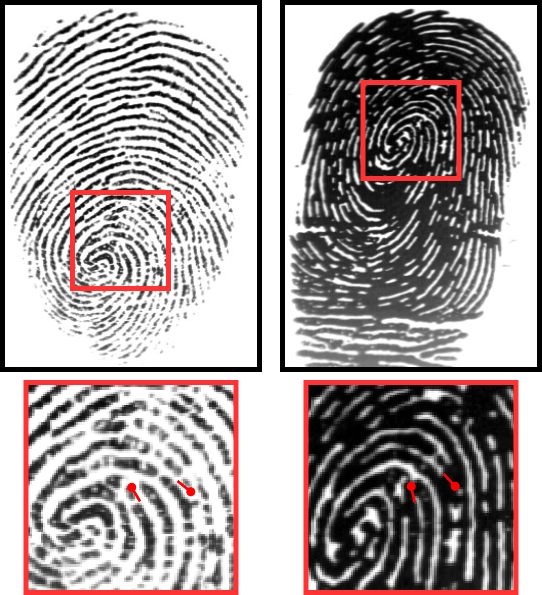}
        \caption{}
        \label{fig:fails_la}
    \end{subfigure}
    \caption{Mutually exclusive failure cases of \cite{DBLP:journals/pami/EngelsmaCJ21} and \cite{latentafis} respectively where the other method works well.}
    \label{fig:fails}
\end{figure}
\begin{figure*}[t]
    \centering
    \begin{subfigure}[t]{0.6\columnwidth}
        \centering
        \includegraphics[width=\columnwidth]{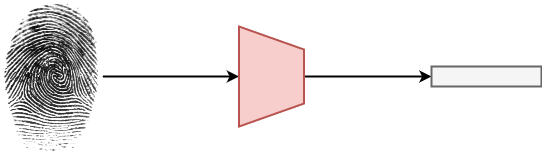}
        \caption{}
        \label{fig:global_intro}
    \end{subfigure}
    \hspace{1em}
    \begin{subfigure}[t]{0.6\columnwidth}
        \centering
        \includegraphics[width=\columnwidth]{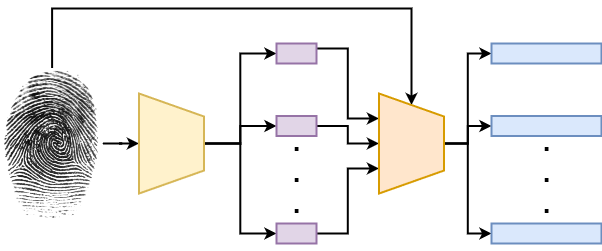}
        \caption{}
        \label{fig:local_intro}
    \end{subfigure}
    \hspace{1em}
    \begin{subfigure}[t]{0.6\columnwidth}
        \centering
        \includegraphics[width=\columnwidth]{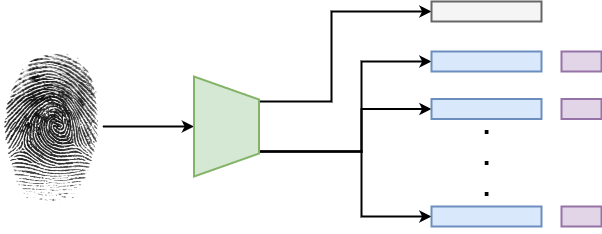}
        \caption{}
        \label{fig:our_intro}
    \end{subfigure}
    \begin{subfigure}[t]{2\columnwidth}
        \centering
        \includegraphics[width=\columnwidth]{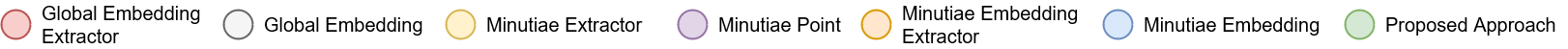}
        \label{fig:legend_intro}
    \end{subfigure}
    \vspace{-3.8mm}
    \caption{Approaches for Fingerprint representation extraction: (a) A global approach that outputs a fixed-length representation. (b) A local approach that first extracts a variable number of minutiae points and then extracts representations corresponding to each of them. (c) Proposed methodology extracts a global fixed length representation, minutiae points and representations corresponding to all  minutiae points in an end-to-end fashion.}
    \label{fig:intro}
    \vspace{-3mm}
\end{figure*}
\\
Hence we posit that using both global and local approaches together can help eliminate a lot of failure cases and consequentially boost performance. But a naive combination poses issues including:
\begin{enumerate*}
    \item Slow feature extraction of \cite{latentafis} due to expensive minutiae detection and multiple feature extraction steps.
    \item Huge memory consumption of \cite{DBLP:journals/pami/EngelsmaCJ21} due to use of big models.
    \item Expensive graph matching of \cite{latentafis} as it evaluates all possible minutiae pairs between two images.
\end{enumerate*}
}
\\
To facilitate the semantic fusion of global and local approaches while avoiding any of the aforementioned issues, we propose a novel methodology as shown in Figure \ref{fig:our_intro}. We can list out the contributions of our paper as:
\begin{itemize}
    \item A novel architecture which, provided a fingerprint image, can in a single forward pass generate:
    \begin{enumerate*}
        \item a global representation
        \item a set of minutiae points
        \item a local representation for each of the predicted minutiae points
    \end{enumerate*}.
    \\
    Whereas other state-of-the-art methods usually use a separate model for each of those steps.
    \item A methodology that:
    \begin{itemize}
        \item is 57.93\% smaller in size than  \cite{DBLP:journals/pami/EngelsmaCJ21}.
        \item does 54.41\% faster feature extraction and 78.47\% faster feature matching as compared to \cite{latentafis}
        \item obtains a state-of-the-art $FRR=\% @ FAR = 0.1\%$ of 1.45 over 6 standard matching databases as compared to 5.65 of \cite{DBLP:journals/pami/EngelsmaCJ21} and 2.32 of \cite{latentafis}.
    \end{itemize}
    \item First work to the best of our knowledge which makes use of a transformer model for fingerprint feature extraction.
\end{itemize}

Our approach makes use of a Convolutional Transformer based architecture because \begin{enumerate*}
    \item the image classification capability of transformer models, as demonstrated in \cite{vit}, make them suitable for global feature extraction.
    \item the ability of transformers to deal with sequences and object localisation, as demonstrated in \cite{detr}, make them suitable for multiple local feature extraction.
    \item it has been shown by \cite{DBLP:conf/icb/CaoJ17}, \cite{DBLP:journals/pami/EngelsmaCJ21}, \cite{SONG2019397}, \cite{Li2019LearningGF} \cite{latentafis} that convolutional networks are effective in fingerprint representation extraction.
    \item the utility of transformers for dealing with fingerprint images has been shown in \cite{trans_fpad}.
\end{enumerate*}

\section{Methodology}
\label{sec:method}
We will first describe the architecture of the model used in our approach, followed by details of training and inference.

\subsection{Architecture}
\label{sec:arch}
\begin{figure}[h]
    \centering
    \includegraphics[width=\columnwidth]{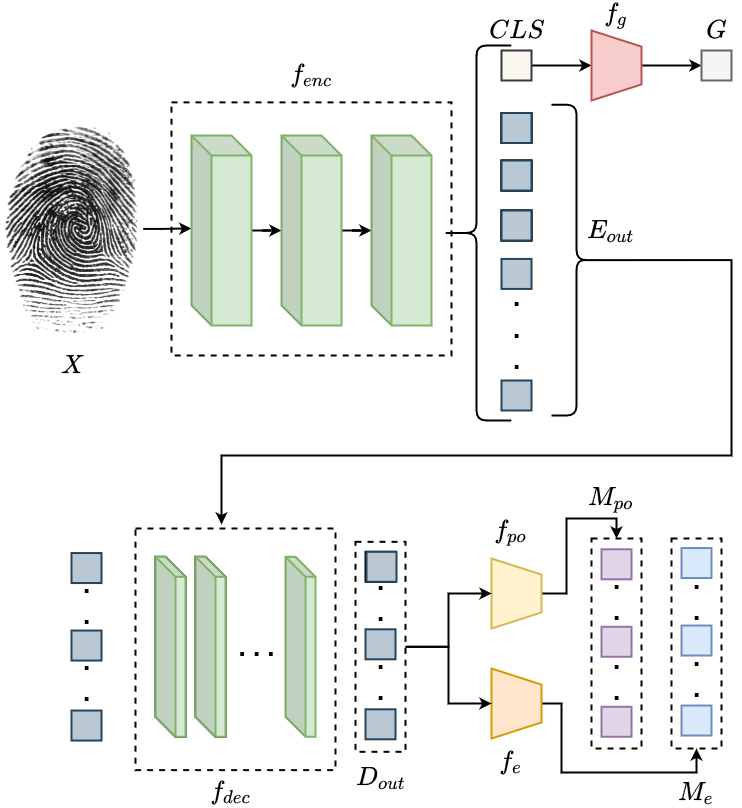}
    \caption{Architecture of the encoder-decoder style model ($f_{model}$) used in our methodology.}
    \label{fig:arch}
    \vspace{-4mm}
\end{figure}

We use an encoder-decoder style architecture, as shown in Figure \ref{fig:arch}, to ``translate'' an input fingerprint image into the required representations. Let the input be a fingerprint image $X \subseteq R^{C\times H\times W}$ where $C$ is the number of channels and $H, W$ represent the size of the image.
\subsubsection{Encoder}
We use a CvT-13 \cite{cvt} model $f_{enc}$  as our encoder which consists of 3 stages of Convolutional Transformer Blocks. On passing X through $f_{enc}$ we obtain a $CLS \subseteq R^{d_{enc}}$ token along with $E_{out} \subseteq R^{L_{enc} \times d_{enc}}$ which is a sequence of $L_{enc}$ embeddings of size $d_{enc}$. Similar to traditional classification tasks we employ the \textit{CLS} token to capture a global essence of the input and hence it is passed through a linear layer $f_g$ to produce the final global representation of our approach $G \subseteq R^{d_g}$ where $d_g$ is the dimensionality of said representation.\\
$E_{out}$ tokens are usually ignored for classification tasks, but similar to \cite{Ahmed2021SiTSV}, \cite{He2021TransFGAT} we posit that these tokens can be used to extract useful auxiliary information about the input. Hence these tokens act as a ``summary'' of the input image and are passed on to the decoder as its ``memory'' input.
\subsubsection{Decoder}
We design a 6 layer decoder $f_{dec}$ where we consider $E_{out}$ as the ``memory'' input and use a placeholder ``target'' sequence of length $L$ similar to the object queries defined by \cite{detr}. $f_{dec}$ computes token embeddings $D_{out} \subseteq R^{L \times d_{dec}}$ where $d_{dec}$ refers to the dimensionality of the decoder. Each of the $L$ tokens of $D_{out}$ is representative of one minutia of the fingerprint and hence $D_{out}$ is passed through:
\begin{itemize}
    \item Pos-Ori MLP ($f_{po})$: This 3 layer MLP outputs $M_{po} \subseteq R^{L \times 3}$ which represents the 2D coordinates and orientation of the $L$ minutiae points predicted in the image.
    \item MinEmb MLP ($f_e$): This 1 layer MLP outputs $M_e \subseteq R^{L \times d_{m}}$ which is the local representation of the $L$ minutiae points of the image as characterized by $M_{po}$. Here $d_m$ represents the dimensionality of the local minutiae representations.
\end{itemize}
Together $M_{po}$ and $M_e$ form the final local representation of the input fingerprint. Here $L$ is a hyperparameter determined by the average number of minutiae in the training data.\\

\begin{algorithm}
\caption{Forward pass of the proposed model. $(f_{model})$}\label{alg:forward}
\begin{algorithmic}[1]
\Require $X \subseteq R^{C \times H \times W}$
\Ensure $G, M_{po}, M_e$
\State $CLS, E_{out} \gets f_{enc}(X)$
\State $G \gets f_g(CLS)$
\State $D_{out} \gets f_{dec}(E_{out})$
\State $M_{po} \gets f_{po}(D_{out})$
\State $M_{e} \gets f_{e}(D_{out})$
\end{algorithmic}
\end{algorithm}


\subsection{Training Stage}
\label{sec:train}
\begin{figure}[t]
    \centering
    \begin{subfigure}{0.18\textwidth}    
        \centering
        \includegraphics[width=\columnwidth]{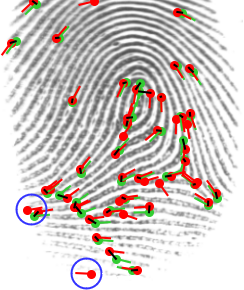}
    \end{subfigure}
    \hspace{1em}
    \begin{subfigure}{0.18\textwidth}    
        \centering
        \includegraphics[width=\columnwidth]{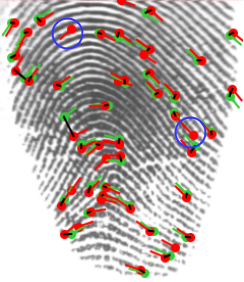}
    \end{subfigure}
    \caption{Green represents the ground truth Verifinger minutiae, whereas red represents the minutiae predicted by our model. Black lines represents the correspondences established by the Hungarian Matcher\cite{Kuhn55thehungarian}. 
    {
    Blue represents spurious minutiae.
    }
    }
    \label{fig:mnt}
    \vspace{-3mm}
\end{figure}

We pass the input through the model as described in Algorithm \ref{alg:forward} and train using a multi-task learning framework so as to facilitate the learning of both global and local perspectives. 
\\
Global Loss $L_{g}$ adopts a teacher-student framework for training the model to generate good global representations for fingerprints. We use \cite{DBLP:journals/pami/EngelsmaCJ21} as our teacher model and use it to generate ground truth embeddings($G' \subseteq R^{192} $) for our training images.
\begin{equation} \label{L_g}
    L_g = MSE(G, G')  
\end{equation}
We use a COTS minutiae extractor called Verifinger to generate ground truth locations and orientations ($M_{po}' \subseteq R^{L\times3}$) of minutiae points for each image in our training dataset. Since our model predicts a fixed number ($L$) of minutiae per image, we pick the top $L$ candidates based on confidence scores. Next, we obtain the ground truth of the local representations corresponding to the minutiae in $M_{po}'$ similar to \cite{popli2021unified}. This gives us ground truth local representations $M_e' \subseteq R^{L\times64}$.
\\
To establish correspondences between the predicted and ground truth minutiae points in an image (Figure \ref{fig:mnt}), we make use of a Hungarian matcher\cite{Kuhn55thehungarian} that scores each correspondence using a linear combination of the L2 distances between the minutiae locations, orientations and representations. Once these correspondences are predicted, we reorder $M_{po}'$ and $M_e'$ to $M_{po}''$ and $M_e''$ respectively and use: 
    \begin{equation} \label{L_po}
    L_{po} = MSE(M_{po}, M_{po}'') 
    \end{equation}
    \begin{equation} \label{L_e}
    L_{e} = MSE(M_{e}, M_{e}'')     
    \end{equation}
We also include supervision over the outputs of the intermediate layers of the decoder. For this we use Eqn. \ref{L_po inter} and \ref{L_e inter} where $M_{po_i}$ and $M_{e_i}$ are the outputs of $f_{po}$ and $f_e$ respectively, when applied on the output of the decoder's $i^{th}$ layer. $M_{po_i}''$ and $M_{e_i}''$ are the re-orderings of the ground truths ($M_{po}'$ and $M_e'$) based on Hungarian matching with $M_{po_i}$ and $M_{e_i}$.
\begin{equation} \label{L_po inter}
    L_{po}^{inter} = \sum_{i=1}^{5} MSE(M_{po_{i}}, M_{po_{i}}'')
\end{equation}
\begin{equation} \label{L_e inter}
    L_{e}^{inter} = \sum_{i=1}^{5} MSE(M_{e_{i}}, M_{e_{i}}'')
\end{equation}

Overall this multi-task, teacher-student training paradigm is guided using the following loss function:
\begin{equation} \label{L_tot}
\begin{aligned}
L_{tot} = \lambda_{g}L_g + \lambda_{po}L_{po} + \lambda_{e}L_e + \\ 
\lambda_{po}^{inter}L_{po}^{inter} + \lambda_{e}^{inter}L_e^{inter} 
\end{aligned}
\end{equation}

\vspace{2mm}
\subsection{Inference Algorithm}
\label{sec:inference}
Given two images, the inference process of our approach has multiple steps as shown in Algorithm \ref{alg:inference}.
\\
\textbf{Feature extraction} stage is executed in steps 1 and 2 which simply includes running the forward pass of the model.
\\
\textbf{Global Matching} (step 3) stage involves a dot product between $G_a$ and $G_b$ as they are fixed-length global representations.
\\
\begin{algorithm}
\caption{Inference using proposed methodology.}\label{alg:inference}
\begin{algorithmic}[1]
\Require $X_a, X_b \subseteq R^{C \times H \times W}$
\Ensure $s$
\State $G_a, M_{poa}, M_{ea} \gets f_{model}(X_a)$
\State $G_b, M_{pob}, M_{eb} \gets f_{model}(X_b)$
\State $s_g \gets Norm(G_a \bullet G_b$) \Comment{Global matching score}
\If{$s_g > \theta_t$} 
    \State $s_l \gets 1$ 
\ElsIf{$s_g < \theta_f$}
    \State $s_l \gets 0$ 
\Else
    \State $s_l \gets Norm(MinMatch(M_{poa}, M_{ea}, M_{pob}, M_{eb}))$ 
    \\\Comment{Local matching score}
\EndIf
\State $s \gets (s_g + s_l) / 2$    \Comment{Final matching score}
\end{algorithmic}
\end{algorithm}

\begin{figure}[t]
    \centering
    \begin{subfigure}{0.24\textwidth}    
        \centering
        \includegraphics[width=\columnwidth]{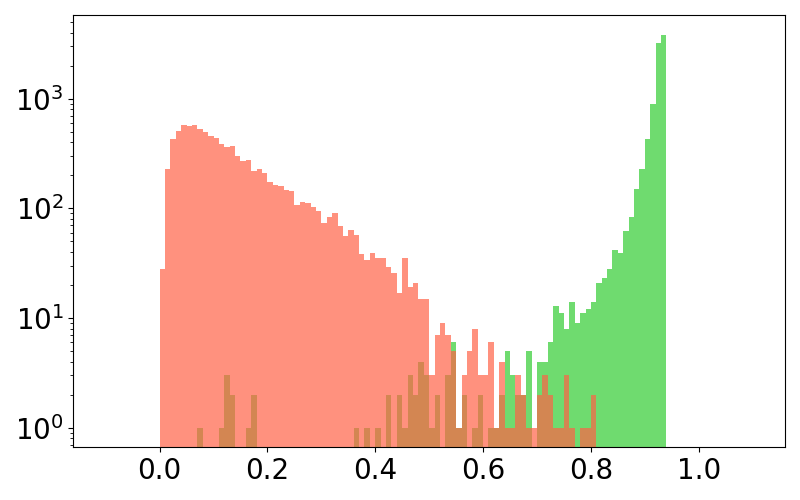}
        \vspace{-0.7cm}
        \caption{}
        \label{fig:scores_global}
    \end{subfigure}
    \begin{subfigure}{0.24\textwidth}    
        \centering
        \includegraphics[width=\columnwidth]{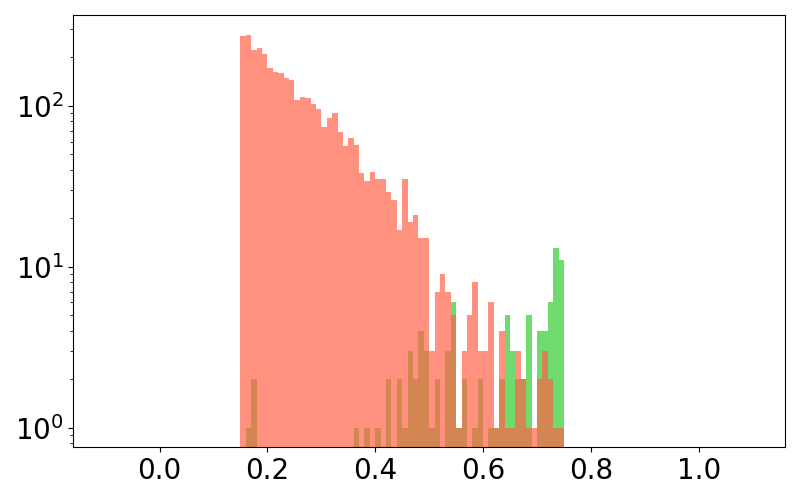}
        \vspace{-0.7cm}
        \caption{}
        \label{fig:scores_selected_global}
    \end{subfigure}
    \begin{subfigure}{0.24\textwidth}    
        \centering
        \includegraphics[width=\columnwidth]{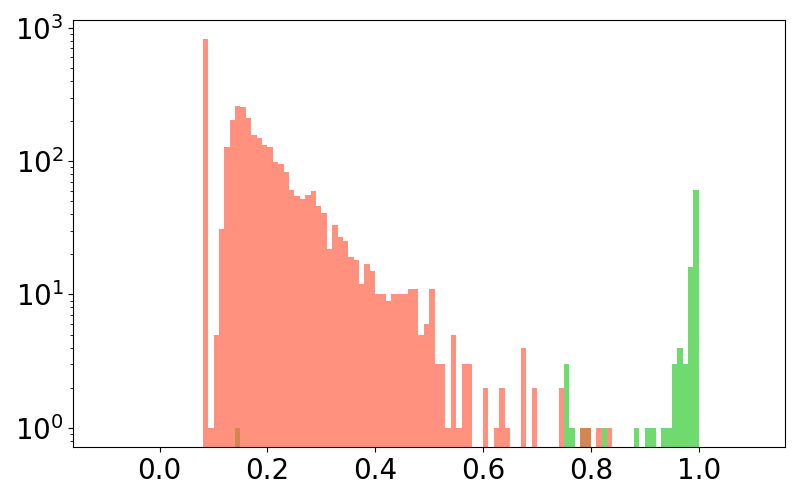}
        \vspace{-0.7cm}
        \caption{}
        \label{fig:scores_selected_local}
    \end{subfigure}
    \begin{subfigure}{0.24\textwidth}    
        \centering
        \includegraphics[width=\columnwidth]{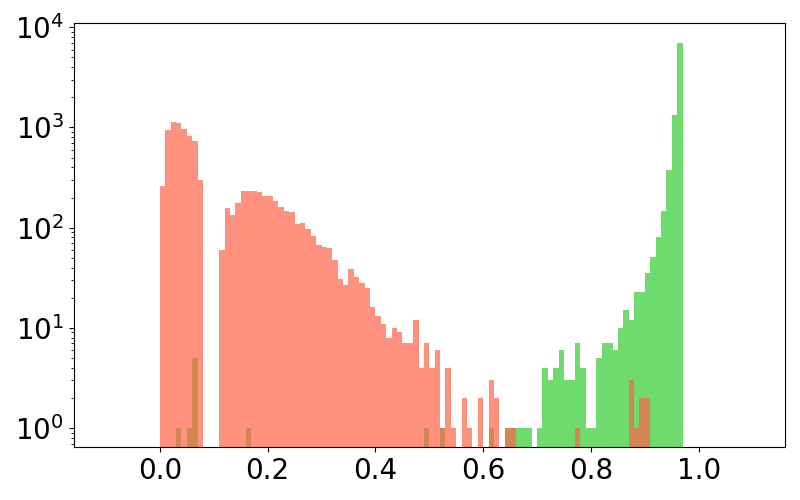}
        \vspace{-0.7cm}
        \caption{}
        \label{fig:scores_masked_avg}
    \end{subfigure}
    \vspace{-2mm}
    \caption{Red color represents scores of impostor pairs and green represents scores of genuine pairs.
    (a) Global Matching Scores for all pairs. 
    (b) Global Matching Scores corresponding to cases where Thresholding permits Local Matching $(\theta_t=0.75, \theta_f=0.15)$ . 
    (c) Local Matching Scores corresponding to cases where Thresholding permits Local Matching.
    (d) Final Matching Scores for all pairs.}
    \label{fig:scores}
    \vspace{-3mm}
\end{figure}

Although our transformer based approach is able to parallelly extract global and local feature representations, to reap the benefits of all that information we need to perform global and local matching both. Due to the expensive nature of minutiae matchers($MinMatch$), this will lead to an inefficient inference algorithm. To counter this we introduce a \textbf{Thresholding} stage which determines the cases where local matching is not expected to add significant value to the approach. Since the calculation of $s_g$ is pretty fast, Thresholding uses it to filter out the cases where solely the global representation leads to a confident inference. Specifically, step 4 represents a highly confident genuine pair prediction and step 6 represents a highly confident impostor pair prediction. In such cases we only use $s_g$ to make the final inference by assigning trivial constants to $s_l$ (steps 5 and 7).
\\
On the other hand if $\theta_f \leq s_g \leq \theta_t$ , we assume that the global representation is not good enough for inference and hence only in these cases Thresholding permits the execution of \textbf{Local Matching} to obtain $s_l$ (step 9).
\\
{
Dot products used for Global Matching produce scores in the range $[0, 1]$, whereas $MinMatch$ used of Local Matching produces unbounded scores greater than 0 (sum of cosine similarities of all the matched minutiae). On combining the raw scores, local matching will dominate and any possible advantages of the global perspective would fade out. Therefore, before combining these scores we need to align their distributions using a \textbf{Score Normalization} function ($Norm$). We make use of the \textit{Double Sigmoid}\cite{bisigm} for this purpose as it is able to bring both distributions in the range $[0,1]$ while being robust and highly efficient\cite{JAIN20052270}.
\\
The final scores are obtained using \textbf{Mean-Fusion} (step 12) to combine the normalized scores. Other score normalization and fusion approaches have been explored in Section \ref{sec:score_norm}.
\\
}
Any inference hyperparameters can be obtained using a hold-out dataset readily available in most real-world scenarios. Score distributions across various steps when Algorithm \ref{alg:inference} is operated on all pairs of a dataset are shown in Figure \ref{fig:scores}. The utility of fusion is clearly represented by the reduced overlapping regions in Figure \ref{fig:scores_masked_avg} as compared  to those in Figure \ref{fig:scores_global}.

\begin{table*}[t]
\centering
\caption{Comparison of the matching performance {($FRR \% @ FAR = 0.1\%$)} on FVC datasets' optical sensors}
\begin{tabular}{  P{2cm} P{1cm} P{1cm} P{1cm} P{1cm} P{1cm} P{1cm} P{1cm} P{1cm}}
        \toprule
        
        \multirow{2}[5]{2cm}{\centering\textbf{Method}} &
        \textbf{2006} & \multicolumn{2}{P{2cm}}{\textbf{2004}} & \multicolumn{2}{P{2cm}}{\textbf{2002}} & \multicolumn{2}{P{2cm}}{\textbf{2000}} & \multirow{2}[5]{1cm}{\centering\textbf{Average\textsuperscript{$\dagger$}}}\\
        
        \cmidrule(r){2-2}\cmidrule(r){3-4}\cmidrule(r){5-6}\cmidrule(r){7-8}
        
        & \textbf{DB2A} & \textbf{DB1A} & \textbf{DB2A} & \textbf{DB1A} & \textbf{DB2A} & \textbf{DB1A} & \textbf{DB3A} \\
        
        \midrule\midrule
        DeepPrint \cite{DBLP:journals/pami/EngelsmaCJ21} & 0.30 & 3.86 & 9.96 & 7.25 & 3.75 & 3.96 & 5.14 & 5.65\\
     
        \midrule
        
        LatentAFIS \cite{latentafis} & \textbf{0.00} & 5.54 & 4.02 & \textbf{0.79} & \textbf{0.36} & 1.39 & 1.79 & 2.32 \\
        
        \midrule
        
        Cascade 
        & \underline{0.01} & \underline{3.39} & \textbf{2.62} & 1.65 & 0.93 & \underline{1.32} & \underline{1.50} & \underline{1.90} \\
        
        \midrule
        
        Our & 0.12 & \textbf{2.00} & \underline{2.73} & \underline{1.11} & \underline{0.86} & \textbf{1.04} & \textbf{0.93} & \textbf{1.45} \\

        \bottomrule
\end{tabular}\\[0.1cm]
{\footnotesize{
Best and second best results are in bold and underline respectively.\\
\textsuperscript{$\dagger$} FVC 2006 DB2A has not been included in the average as it is used as a hold-out dataset in Sections \ref{inference_parameters} and \ref{sec:score_norm}. 
}}
\label{table: matching}
\vspace{-3mm}
\end{table*}
\section{Datasets}
\label{sec:data}
{
For training and validation purposes we have used an in-house dataset made up of fingerprint images captured using CrossMatch and DigitalPersona optical sensors. The training and validation splits consist of 130K and 38K images respectively. 
}
For evaluation of fingerprint matching we report on different sensors of FVC 2000 \cite{fvc2000}, FVC 2002 \cite{fvc2002}, FVC 2004 \cite{fvc2004} and FVC 2006 \cite{fvc2006} datasets following the official protocols. Each sensor of the FVC 2000, 2002 and 2004 datasets contains 100 subjects with 8 impressions per subject, leading to 2800 genuine 
($=100 \times \Comb{8}{2}$) 
and 4950 impostor 
($ =\Comb{100}{2}$) 
comparisons, while each sensor within the FVC 2006 dataset contains 140 subjects with 12 impressions per subject leading to 9240 
($ = 140 \times \Comb{12}{2}$) 
genuine and 9730 
($ = \Comb{140}{2}$) 
impostor comparisons.\\

\section{Experiments and Results}
\label{sec:exp}
We compare our proposed methodology with \cite{DBLP:journals/pami/EngelsmaCJ21} and \cite{latentafis} as they are the current state-of-the-art approaches for global and local matching respectively. We also compare with the performance obtained by using a straightforward combination of \cite{DBLP:journals/pami/EngelsmaCJ21} and \cite{latentafis} called \textbf{\textit{Cascade}}. In this, feature extraction and matching are performed by running both approaches independently to completion and fusing the final matching scores as described in Algorithm \ref{alg:inference}. This is done to provide insight into how our methodology helps in parallelizing aspects of global and local approaches.  We do not make comparisons with commercial matchers like VeriFinger as their proprietary nature prevents the evaluation of their complexity and we want to focus on generating an efficient approach. 
\\
{
Since we train using a private dataset, to make our work reproducible we present the experimentation and details of a model trained using public datasets in the supplementary.
}

\subsection{Matching}
\label{sec:matching}
Since our training dataset consists of optical images, in this experiment we evaluate on the various optical sensors of FVC datasets to assess the same-sensor performance of our approach.
\\
The minutiae predicted by our methodology obtain a Goodness Index and average positional error\cite{benchmin} of 0.31 and 5.51 pixels respectively (with a distance threshold of 20 pixels) which is comparable to 0.20 and 5.29 pixels obtained by the open source minutiae extractor \cite{mindtct}. This indicates the utility of our approach for the task of minutiae detection. \\
In Table \ref{table: matching} we report the metric $FRR \% @ FAR = 0.1\%$ 
and the ROC curves for \cite{fvc2004} can be found in the supplementary. 
Our methodology clearly beats both of the current state-of-the-art methods on 4 and the Cascade baseline on 5 datasets individually. Moreover we beat all three baselines on an average of 6 datasets. Hence this proves the state-of-the-art nature of our approach for the task of matching.
\subsection{Time and Memory}
\label{section:time_and_mem}
\begin{figure*}[t]
    \centering
    \begin{subfigure}{0.32\textwidth}    
        \centering
        \includegraphics[width=\columnwidth]{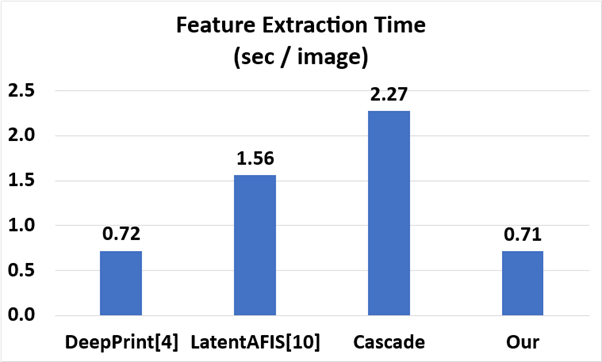}
        \caption{}
        \label{fig:time_ext}
    \end{subfigure}
    \begin{subfigure}{0.32\textwidth}    
        \centering
        \includegraphics[width=\columnwidth]{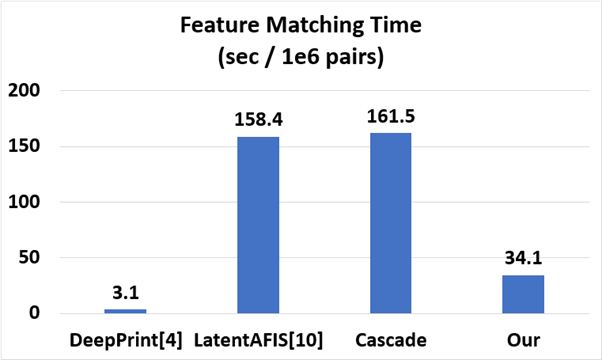}
        \caption{}
        \label{fig:time_match}
    \end{subfigure}
    \begin{subfigure}{0.32\textwidth}    
        \centering
        \includegraphics[width=\columnwidth]{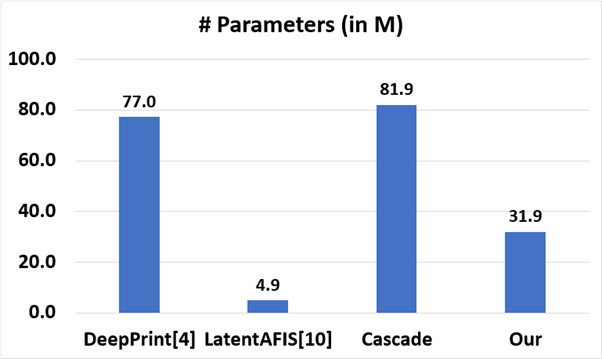}
        \caption{}
        \label{fig:params}
    \end{subfigure}
    \vspace{-1mm}
    \caption{Comparison of time and memory requirements with state-of-the-art methods.}
    \label{fig:time_and_mem}
    \vspace{-4mm}
\end{figure*}

We can observe in Figure \ref{fig:time_ext} that our method is 54.41\% faster than \cite{latentafis} and has almost the same speed as \cite{DBLP:journals/pami/EngelsmaCJ21} in the feature extraction stage. This leads to a 68.72\% speed-up as compared to Cascade for the same.
\\
Figure \ref{fig:time_match} shows that our method achieves a significant speed-up in the feature matching stage as compared to \cite{latentafis} which leads to a 78.88\% speed-up as compared to Cascade.
\\
Finally, Figure \ref{fig:params} shows that our model is 57.93\% smaller than the model used in \cite{DBLP:journals/pami/EngelsmaCJ21}, which is a significant advantage to compensate for the difference in matching time of the two approaches. Moreover, it leads to 61.05\% more memory efficiency as compared to Cascade. Although we show that our model contains more parameters than \cite{latentafis}, it is important to note that here we have excluded the parameters required by \cite{latentafis} to extract minutiae during inference. Verifinger, a black-box operator, is used for that purpose and hence its complexity cannot be evaluated. Whereas our model includes an in-built minutiae detector.\\
This shows that our methodology is able to bring together the semantic advantages of the two state-of-the-art approaches more efficiently than Cascade by improving upon the bottleneck factors including the time taken by \cite{latentafis} and the memory consumption of \cite{DBLP:journals/pami/EngelsmaCJ21}.

\subsection{{Generalization}}
\label{sec:generalization}
{
To evaluate the cross-sensor generalization performance of our approach we evaluate on the non-optical sensors of FVC datasets as shown in Table \ref{table: generalization}. Note that our training data consists of only optical images and hence thermal, synthetic and capacitive sensors are unseen domains for our model. 
We observe that our approach is able to obtain state-of-the-art performance across three sensors belonging to different domains.
}
\begin{table}[h]
\centering
\caption{Comparison of the matching performance ($FRR \% @ FAR = 1.0\%$) on FVC datasets' non-optical sensors}
\begin{tabular}{  P{1.65cm} P{1.1cm} P{1.1cm} P{1.1cm} P{1cm} }
        \toprule
        
        \multirow{2}[5]{1.65cm}{\centering\textbf{Method}} &
        \textbf{Thermal} & \textbf{Synthetic} & \textbf{Capacitive} & \multirow{2}[5]{1cm}{\centering\textbf{Average}}\\
        
        \cmidrule(r){2-4}
        
        & \textbf{2006DB3A} & \textbf{2006DB4A} & \textbf{2000DB2A} &\\
        
        \midrule\midrule
        DeepPrint\cite{DBLP:journals/pami/EngelsmaCJ21} & 5.17 & 14.58 & 1.14 & 6.96
        \\
        
        \midrule
        
        LatentAfis\cite{latentafis} & 1.20 & 2.42 & 0.14 & 1.25\\
        
        \midrule
       
        Cascade & 1.01 & 1.63 & 0.11 & 0.92  \\
        
        \midrule
        
        Our & 1.27 & 1.18 & 0.11 & 0.85 \\
        
        \bottomrule
\end{tabular}
\label{table: generalization}
\vspace{-3mm}
\end{table}

\section{Ablation Study}

\subsection{Loss function}
In this study we try to play around with the supervision provided to train the model by changing the values of the hyperparameters in Eqn. \ref{L_tot} and observe the differences in performance on FVC 2004 DB1A\cite{fvc2004}. The magnitude of these hyperparameters was chosen based on the magnitude of loss values on the validation set. \\
It is clear that a model trained with a combination of local and global supervision is superior to a model trained solely for one of them. Moreover we also observe a significant benefit of including the intermediate loss terms.
\begin{table}[h]
\centering
\caption{$FRR\%@FAR=0.1\%$ for FVC 2004 DB1A\cite{fvc2004} using various loss functions.}
{\renewcommand{\arraystretch}{1}%
\begin{tabular}{ P{0.7cm} P{0.7cm} P{0.7cm} P{0.7cm} P{0.7cm} P{1.5cm}}
        \toprule
        $\lambda_{g}$ & $\lambda_{po}$  & $\lambda_{e}$ & $\lambda_{po}^{inter}$  & $\lambda_{e}^{inter}$ & $FRR\%$ \\
        
        \midrule\midrule
        1 & 0 & 0 & 0 & 0 & 5.32
        \\
        \midrule
        0 & 1 & 1 & 0 & 0 & 6.89
        \\
        \midrule
        10 & 1 & 1 & 0 & 0 & 2.79
        \\
        \midrule
        60 & 1 & 1 & 1 & 1 & 2.00
        \\
        \bottomrule
        
\end{tabular}}
\flushleft{\footnotesize{}}
\label{table: loss_ablation}
\vspace{-3mm}
\end{table}

\subsection{Inference parameters} \label{inference_parameters}
In Algorithm \ref{alg:inference} we have made use of $\theta_t$ and $\theta_f$ for Thresholding. In practice, their values can be obtained using some hold-out representative dataset. Here we make use of the FVC 2006 DB2A\cite{fvc2006} dataset to experiment with different combinations of values for these hyperparameters. \\
Our methodology retains all of its utility except for feature matching speed-up when trivial constants are used for $\theta_t$ and $\theta_f$. Such a case is shown in the first row of Table \ref{table: inference_ablation}. Subsequently we observe that as the gap between $\theta_t$ and $\theta_f$ decreases, the matching time decreases too, as the number of cases for which Thresholding permits local matching decreases. But after a while it leads to a decrease in performance as the advantage of fusing local scores with global ones fades out.  For our results presented in Section \ref{sec:exp} we have used $\theta_t=0.75$ and $\theta_f=0.15$ as this provides the best trade-off between matching time and performance as shown in Table \ref{table: inference_ablation}.\\

\begin{table}[h]
\vspace{-2mm}
\centering
\caption{$FRR\%@FAR=0.1\%$ for FVC 2006 DB2A\cite{fvc2006} using various Thresholding hyperparameter values.}
{\renewcommand{\arraystretch}{1}%
\begin{tabular}{ P{0.7cm} P{0.7cm} P{1.7cm} P{2.5cm} P{1cm}}
        \toprule
        
       \textbf{$\theta_t$} & \textbf{$\theta_f$} & \textbf{\%times when $\theta_f \leq s_g \leq \theta_t$} & \textbf{Matching Time \quad(sec / 1e6 pairs)} & $FRR\%$ \\
        
        \midrule\midrule
        $>1$ & $<0$  & 100.00 & 161.48 & 0.11 \\
        \midrule
        $0.8$ & $0.1$ & 25.03  & 42.74 & 0.11 \\
        \midrule
        $0.75$ & $0.15$ & 19.56  & 34.08 & 0.12 \\
        \midrule
        $0.7$ & $0.2$ & 12.97  & 23.63 & 0.83 \\
        \bottomrule
        
\end{tabular}}
\flushleft{\footnotesize{}}
\label{table: inference_ablation}
\vspace{-2mm}
\end{table}

\subsection{{Score Normalization and Fusion}}
\label{sec:score_norm}
{
In Table \ref{table: score_norm} we set $\theta_t>1$ and $\theta_f<0$ so that we can explore the various score normalization and fusion techniques\cite{JAIN20052270} independent of Thresholding. We again use FVC2006 DB2A\cite{fvc2006} as a hold-out dataset to determine any hyperparameters required for normalization.
\\
Double Sigmoid\cite{bisigm}, being the most robust and efficient technique, gives the best performance among normalizations. Although Max-Fusion gives a better performance with Double Sigmoid\cite{bisigm} than Mean-Fusion, it is much more sensitive to normalization and Thresholding.
}
\begin{table}[h]
\centering
\caption{Average $FRR\%@FAR=0.1\%$ over optical FVC datasets(as in Table \ref{table: matching}) w.r.t. score normalization and fusion.}
{\renewcommand{\arraystretch}{1}%
\begin{tabular}{ P{2.5cm} P{1.6cm} P{1.5cm} P{1.5cm}}
        \toprule
        
       \textbf{Normalization} & \textbf{Mean-Fusion} & \textbf{Max-Fusion} & \textbf{Min-Fusion} \\
        
        \midrule\midrule
        None & 2.22 & 2.51 & 5.75\\
        \midrule
        Min-max & 1.79 & 5.82 & 2.51\\
        \midrule
        $z$-score & 1.77 & 4.57 & 2.42\\
        \midrule
        Median and MAD & 1.91 & 2.51 & 3.93\\
        \midrule
        Modified tanh\cite{Latha2011EfficientAT} & 1.77 & 4.57 & 2.42\\
        \midrule
        \textbf{Double Sigmoid\cite{bisigm}} & 1.68 & 1.46 & 3.70\\
        \bottomrule
        
\end{tabular}}
\flushleft{\footnotesize{}}
\label{table: score_norm}
\vspace{-1mm}
\end{table}

        
        
        

\subsection{{Amount of Local Information}}
\label{sec:amt_local_info}
{
In this study we set $\theta_t>1$ and $\theta_f<0$ and alter the amount of local information available for fusing by controlling the number of minutiae points used for Local Matching during inference. Figure \ref{fig:filtermnt} clearly shows that a smaller subset leads to faster matching speeds but worse performances. This introduces a trade-off, but it is interesting to note that the deterioration in performance of our method is significantly less as compared to \cite{latentafis} due to the constant advantage of fusion with global scores. Hence this approach of selecting a random minutiae subset and merging local with global scores in all cases can be used as an alternative to Thresholding to speed-up feature matching while maintaining performance.
\begin{figure}[h]
    \vspace{-2mm}
    \centering
    \includegraphics[width=0.95\columnwidth]{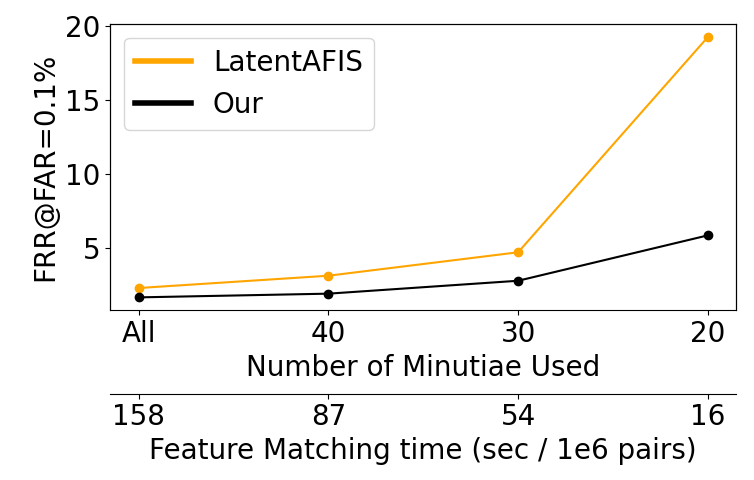}
    \vspace{-1mm}
    \caption{Variation in average performance(as in Table \ref{table: matching}) and feature matching time with respect to number of minutiae used.}
    \label{fig:filtermnt}
    \vspace{-2mm}
\end{figure}
}

\section{Conclusions and Future Work}
\label{sec:conclusion}
We presented a novel methodology for fingerprint feature extraction where a single end-to-end transformer model can be used to parallelly and efficiently obtain global representations, minutiae predictions as well as local representations from fingerprint images instead of employing a separate module for each of them. To prove the utility of combining global and local perspectives, we also present an efficient feature matching process over these representations to achieve state-of-the-art performances on multiple datasets.\\
{
A possible future extension of this work can be incorporating a filtering mechanism for spurious minutiae (as shown in Figure \ref{fig:mnt}) so that we can use different number of minutiae points for different images instead of a constant $L$. This is important as this eliminates false minutiae matches and any potential attacks exploiting a fixed number of local features. We are also working on merging this work with \cite{popli2021unified} to integrate spoof detection in this methodology and eliminate the vulnerability against presentations attacks.
}

\section{Supplementary: Private Dataset Experiments}
This section includes supplementary material related to the experiments performed using models trained on the in-house dataset.
\subsection{ROC curves}
\begin{figure}[h]
    \centering
    \begin{subfigure}{0.45\textwidth}    
        \centering
        \includegraphics[width=\columnwidth]{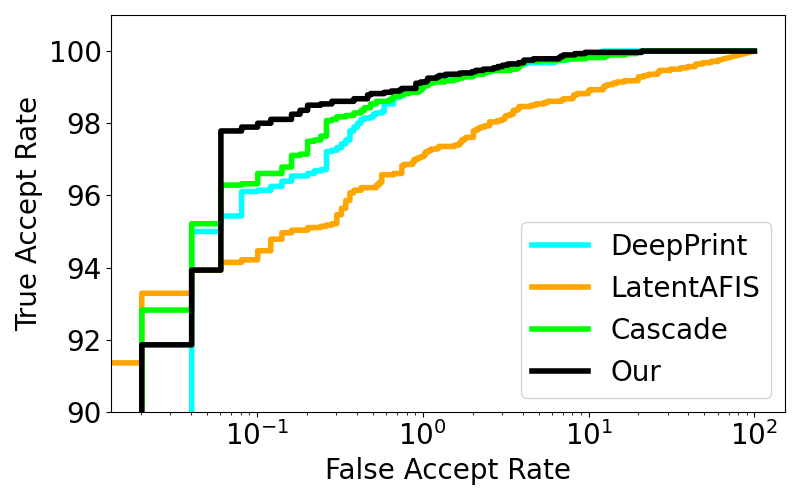}
    \end{subfigure}
    \begin{subfigure}{0.45\textwidth}    
        \centering
        \includegraphics[width=\columnwidth]{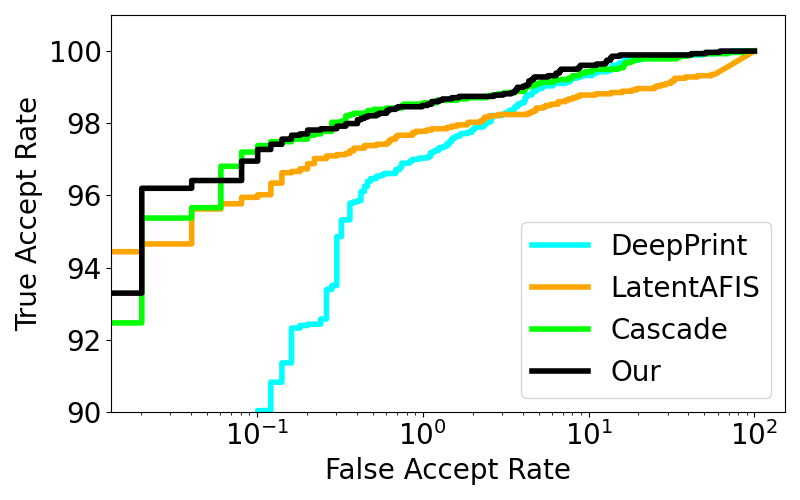}
    \end{subfigure}
    \caption{ROC curves of our method, DeepPrint\cite{DBLP:journals/pami/EngelsmaCJ21} and LatentAfis\cite{latentafis} for FVC2004 DB1A and DB2A respectively.}
    \label{fig:roc}
\end{figure}
Figure \ref{fig:roc} shows the curves of True Accept Rate vs False Accept Rate for FVC2004\cite{fvc2004} DB1A and DB2A. Here True Accept Rate refers to the percentage of genuine pairs predicted as genuine and False Accept Rate refers to the percentage of impostor pairs predicted as genuine. These clearly show the performance advantage of our method at various thresholds as compared to \cite{DBLP:journals/pami/EngelsmaCJ21}, \cite{latentafis} and Cascade.

\subsection{Generalization examples}
\begin{figure}[h]
    \centering
    \begin{subfigure}{\columnwidth}    
        \centering
        \includegraphics[width=0.25\columnwidth]{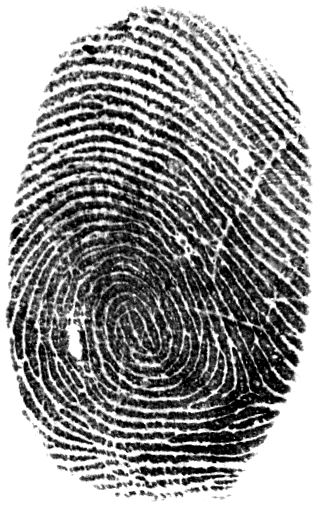}
        \includegraphics[width=0.25\columnwidth]{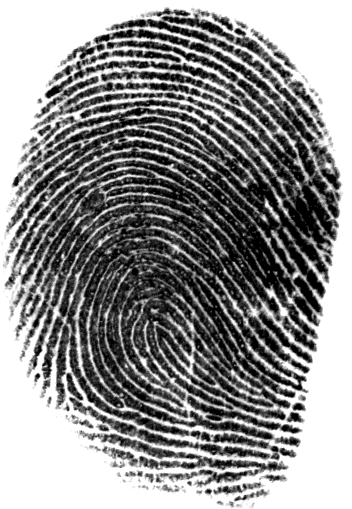}
        \includegraphics[width=0.25\columnwidth]{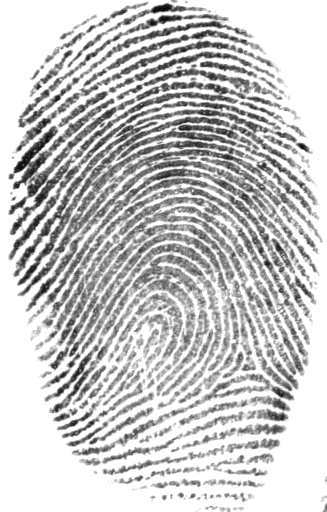}
        \caption{Optical images used for training and same-sensor testing.}
    \end{subfigure}
    \\[0.1cm]
    \begin{subfigure}{\columnwidth}    
        \centering
        \includegraphics[width=0.25\columnwidth]{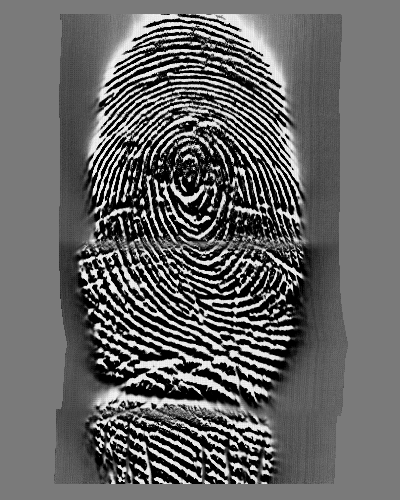}
        \includegraphics[width=0.25\columnwidth]{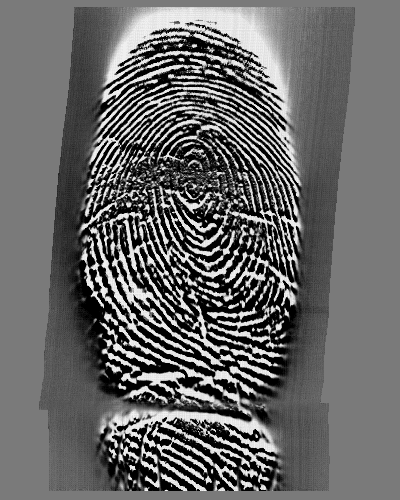}
        \includegraphics[width=0.25\columnwidth]{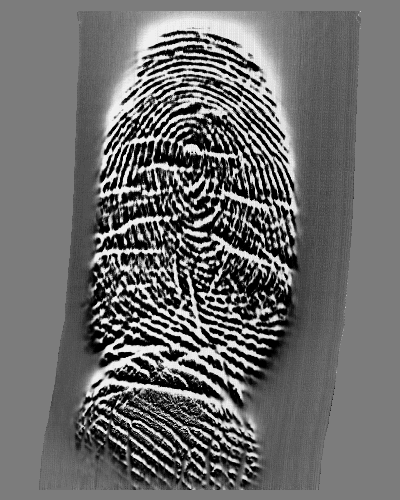}
        \caption{Thermal images.}
    \end{subfigure}
    \\[0.1cm]
    \begin{subfigure}{\columnwidth}    
        \centering
        \includegraphics[width=0.25\columnwidth]{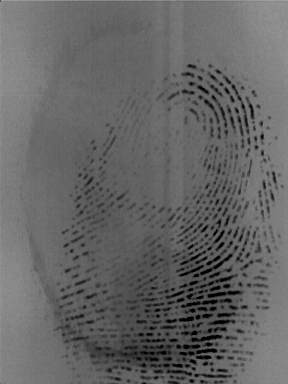}
        \includegraphics[width=0.25\columnwidth]{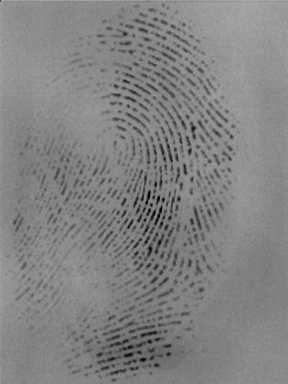}
        \includegraphics[width=0.25\columnwidth]{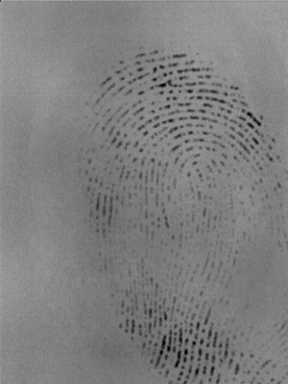}
        \caption{Synthetic images.}
    \end{subfigure}
    \\[0.1cm]
    \begin{subfigure}{\columnwidth}    
        \centering
        \includegraphics[width=0.25\columnwidth]{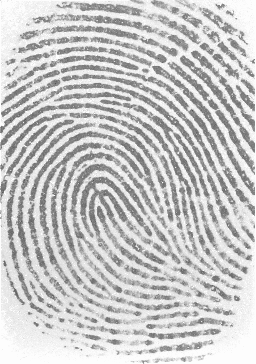}
        \includegraphics[width=0.25\columnwidth]{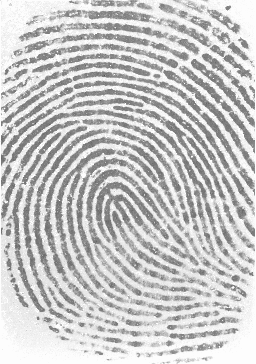}
        \includegraphics[width=0.25\columnwidth]{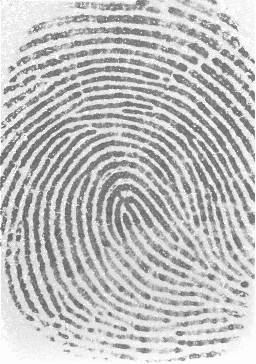}
        \caption{Capacitive images.}
    \end{subfigure}
    \caption{}
    \label{fig:generalization_egs}
\end{figure}

Figure \ref{fig:generalization_egs} shows examples of images from the different kinds of sensors used for experimentation.

\section{Supplementary: Ablation Study of Number of Decoder Layers}
We try to alter the number of layers in $f_{dec}$ and observe the variation in performance. Table \ref{table: declayers_ablation} clearly shows that with increasing number of decoder layers the validation losses $L_{po}$ and $L_e$ converge to better minimas. But with increasing number of decoder layers the model complexity increases and hence the inference time for feature extraction also increases. Our choice of using 6 layers in our main experimentation is justified as we are able to get the best model with feature extraction time per image comparable to that of \cite{DBLP:journals/pami/EngelsmaCJ21}.
\begin{table}[h]
\centering
\caption{Variation in validation losses and inference time with respect to the number of layers in the decoder.}
{\renewcommand{\arraystretch}{1}%
\begin{tabular}{ P{1.5cm} P{1.5cm} P{1.5cm} P{1.5cm}}
        \toprule
        \textbf{\# Decoder Layers} & \textbf{$L_{po} \times 10^{-3}$} & \textbf{$L_e \times 10^{-3}$} & \textbf{CPU Inference Time (sec)}
        \\
        
        \midrule\midrule
        1 & 5.77 & 4.00 & 0.660 \\
        \midrule
        2 & 5.31 & 3.92 & 0.674 \\
        \midrule
        3 & 5.07 & 3.86 & 0.681 \\
        \midrule
        4 & 4.88 & 3.75 & 0.690 \\
        \midrule
        5 & 4.78 & 3.70 & 0.697 \\
        \midrule
        6 & 4.69 & 3.65 & 0.709 \\
        
        \bottomrule
        
\end{tabular}}\\[0.2cm]
\label{table: declayers_ablation}
\end{table}

\section{Public Dataset Experiments}
This section repeats the experimentation performed in the main paper, but here we train our models with public datasets. This is done to make our work accessible and reproducible.

\subsection{Datasets}
For training and validation purposes we use a combination of multiple datasets to maximize the effectiveness of the training stage. They include the plain images from NIST SD 300\cite{sd300}, optical  images from NIST SD 302\cite{sd302}, MCYT\cite{mcyt} and live images from LivDet 2011\cite{ld2011}, 2013\cite{ld2013}, 2015\cite{ld2015} and 2017\cite{ld2017}. This gives us a huge dataset that spans multiple sources and years. The details of each of the aforementioned datasets can be found in Table \ref{table: train_data}.\\
For testing we use the same datasets and procedures as described in the main paper.
\begin{table}[h]
\centering
\footnotesize
\caption{Details of the training and validation dataset.}
{\renewcommand{\arraystretch}{1}%
\begin{tabular}{ P{3cm} P{1.5cm} P{1.5cm}}
        \toprule
        
        \textbf{Name} & \textbf{\# train images} & \textbf{\# val images} \\[0.2cm]
        
        \midrule\midrule
        NIST SD 300 \cite{sd300} & 6927 & 1763\\[0.2cm]
        \midrule
        NIST SD 302 \cite{sd302} & 5332 & 1304 \\[0.2cm]
        \midrule
        MCYT \cite{mcyt} & 19199 & 4800  \\[0.2cm]
        \midrule
        LivDet \cite{ld2011, ld2013, ld2015, ld2017} & 13749 & 15118 \\[0.2cm]
        \midrule\midrule
        Total & $\sim$ 45K & $\sim$ 23K \\
        \bottomrule
        
\end{tabular}}
\flushleft{\footnotesize{}}
\label{table: train_data}
\end{table}

\subsection{Results}
Table \ref{table: matching_public} presents the results obtained by our approach when trained using the public datasets. Since this dataset is smaller than the in-house data we observe some drop in performance, but we are still much better than the \cite{DBLP:journals/pami/EngelsmaCJ21} and \cite{latentafis} baselines. As compared to Cascade, we still retain all the time and memory advantages discussed in the main paper.
\begin{table*}[t]
\centering
\caption{Comparison of the matching performance ($FRR \% @ FAR = 1.0\%$) on FVC datasets' optical sensors}
\begin{tabular}{  P{2cm} P{1cm} P{1cm} P{1cm} P{1cm} P{1cm} P{1cm} P{1cm} P{1cm}}
        \toprule
        
        \multirow{2}[5]{2cm}{\centering\textbf{Method}} &
        \textbf{2006} & \multicolumn{2}{P{2cm}}{\textbf{2004}} & \multicolumn{2}{P{2cm}}{\textbf{2002}} & \multicolumn{2}{P{2cm}}{\textbf{2000}} & \multirow{2}[5]{1cm}{\centering\textbf{Average\textsuperscript{$\dagger$}}}\\
        
        \cmidrule(r){2-2}\cmidrule(r){3-4}\cmidrule(r){5-6}\cmidrule(r){7-8}
        
        & \textbf{DB2A} & \textbf{DB1A} & \textbf{DB2A} & \textbf{DB1A} & \textbf{DB2A} & \textbf{DB1A} & \textbf{DB3A} \\
        
        \midrule\midrule
        DeepPrint \cite{DBLP:journals/pami/EngelsmaCJ21} & 0.12 & 0.96 & 2.96 & 4.11 & 1.82 & 1.79 & 2.04 & 2.28 \\
     
        \midrule
        
        LatentAFIS \cite{latentafis} & 0.00 & 2.89 & 2.22 & 0.54 & 0.25 & 0.68 & 1.25 & 1.31 \\
        
        \midrule
        
        Cascade 
        &  0.00 & 0.96 & 1.44 & 0.43 & 0.21 & 0.25 & 0.43 & 0.62\\
        
        \midrule
        
        Our Public ($\theta_t = 0.75$ and $\theta_f = 0.15$) & 0.08 & 1.21 & 1.61 & 0.75 & 0.93 & 0.39 & 0.50 & 0.90 \\
        
        \bottomrule
\end{tabular}\\[0.1cm]
{\footnotesize{
\textsuperscript{$\dagger$} FVC 2006 DB2A has not been included in the average. 
}}
\label{table: matching_public}
\vspace{-3mm}
\end{table*}

\section{Supplementary: Implementation Details}
We preprocess the images by segmenting them, followed by a center crop / padding to size $441 \times 441$, a resize to size $410 \times 410$ and another center crop of size $384 \times 384$. The coordinates and orientation of the minutiae points are altered accordingly. This procedure is similar to the one followed in \cite{DBLP:journals/pami/EngelsmaCJ21}.
\\
All experimentation was performed using the PyTorch framework and two Nvidia 2080-Ti GPUs. An Intel(R) Xeon(R) CPU E5-2640 v4 @ 2.40GHz was used for inference purposes. A batch size of 32 was used to train all models along with an adamW\cite{loshchilov2018decoupled} optimizer and cosine learning rate schedule of 100 epochs initialised with $1e^{-4}$. The value of L was set to 50 based on the average number of minutiae in the training set. We set $d_{enc}=d_{dec}=384$ and based on the ground truth the values for $d_g$ and $d_m$ were set to 192 and 64 respectively.






%
\bibliographystyle{IEEEtran}
\bibliography{IEEEabrv,IEEEexample}


\end{document}